\documentclass{bmvc2k}

\usepackage{algorithm}
\usepackage{algpseudocode}
\usepackage{multirow}
\usepackage{wrapfig}
\usepackage{bm}

\DeclareMathOperator*{\argmin}{arg\,min}

\title{Learning Human Poses from Actions}

\addauthor{Aditya Arun}{aditya.arun@research.iiit.ac.in}{1}
\addauthor{C.V. Jawahar}{jawahar@iiit.ac.in}{1}
\addauthor{M. Pawan Kumar}{pawan@robots.ox.ac.uk}{2}

\addinstitution{
 IIIT Hyderabad
}
\addinstitution{
 University of Oxford \& \\
 The Alan Turing Institute
}

\runninghead{Aditya, Jawahar, Pawan}{Learning Human Poses from Actions}

\def\etal{\emph{et al}\bmvaOneDot}

\begin{document}

\maketitle

\begin{abstract}
We consider the task of learning to estimate human pose in still images. In order to avoid the high cost of full supervision, we propose to use a diverse data set, which consists of two types of annotations: (i) a small number of images are labeled using the expensive ground-truth pose; and (ii) other images are labeled using the inexpensive action label. As action information helps narrow down the pose of a human, we argue that this approach can help reduce the cost of training without significantly affecting the accuracy. To demonstrate this we design a probabilistic framework that employs two distributions: (i) a conditional distribution to model the uncertainty over the human pose given the image and the action; and (ii) a prediction distribution, which provides the pose of an image without using any action information. We jointly estimate the parameters of the two aforementioned distributions by minimizing their dissimilarity coefficient, as measured by a task-specific loss function. During both training and testing, we only require an efficient sampling strategy for both the aforementioned distributions. This allows us to use deep probabilistic networks that are capable of providing accurate pose estimates for previously unseen images. Using the MPII data set, we show that our approach outperforms baseline methods that either do not use the diverse annotations or rely on pointwise estimates of the pose.
\end{abstract}

\section{Introduction}
\label{sec:introduction}
Current methods for learning human pose estimation from still images require the collection of a fully annotated data set, where each training sample consists of an image of a person, together with its ground-truth joint locations. The collection of such detailed annotations is onerous and expensive, which makes this approach unscalable. We propose to alleviate the deficiency of fully supervised learning by using a diverse data set. Part of the images of the data set are labeled with expensive pose annotations, while the remaining images are labeled with inexpensive action annotations.

Throughout the paper, we assume that the distribution of the images labeled with different types of annotations is the same (which is a necessary assumption for learning) and that the annotations themselves are noise-free. Under these assumptions, we argue that action information can be used to learn pose estimation. Note that earlier works have exploited the relationship between action and pose for action recognition. However, our problem is significantly more challenging due to the high uncertainty in pose given the action. In order to model this uncertainty, we propose to use a probabilistic learning formulation. A typical probabilistic formulation would learn a joint distribution of the pose and the action given an image. In order to make a prediction on a test sample, where action information is not known, it would marginalize over all possible actions. In other words, it would use one set of parameters for two distinct tasks: (i) model the uncertainty in the pose for every action; and (ii) predict the pose given an image.

As our goal is to make an accurate pose prediction, we argue that such an approach would waste the modeling capability of a distribution in representing pose uncertainty in the presence of action information. In other words, the parameters of the distribution will be tuned to perform well in the presence of action information, which will not be available during testing. Instead, we use two different distributions for the two different tasks: (i) a {\em conditional distribution} of the pose given the image and the action; and (ii) a {\em prediction distribution} of the pose given the image. 

\begin{wrapfigure}{r}{0.35\textwidth}
\vspace{-11mm}
\label{fig:motivation}
\begin{center}
   \includegraphics[scale=0.45]{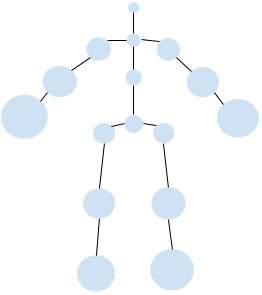}
\end{center}
   \vspace{-3mm}
   \caption{\em Average entropy of joints in test images over a stick figure. The radius of circle around a joint is proportional to the joint's entropy.
   }
   \vspace{-3mm}
\end{wrapfigure}

We jointly estimate the parameters of the two distributions by minimizing their dissimilarity coefficient~\cite{rao1982diversity}, which measures the distance between two distributions using a task-specific loss function. By transferring the information from the conditional distribution to the prediction distribution, we learn to estimate the pose of a human using a diverse data set. Figure~\ref{fig:motivation} shows the necessity of using a probabilistic model. Specifically, the figure shows the average entropy of each joint as predicted by our model on test images. We observe that the most articulate joints like wrists and ankles have highest entropy, which a non probabilistic network does not explicitly model.

While our approach can be used in conjunction with any parametric family of distributions, in this work we focus on the state of the art deep probabilistic networks. Specifically, we model both the conditional and the prediction distributions using a DISCO Net~\cite{bouchacourt2016disco}, which allows us to efficiently sample from the two distributions. As will be seen later, the ability to sample efficiently is sufficient to make both training and testing computationally feasible.

We demonstrate the efficacy of our approach using the publicly available MPII Human Pose data set~\cite{andriluka20142d}. We discard the pose information of a portion of the training samples but retain the action information for all the samples in order to generate a diverse data set. We provide a thorough comparison of our probabilistic approach with two natural baselines. First, a fully supervised approach, which discards the weakly supervised samples that have been labeled using only the action information. Second, a pointwise model that uses a self-paced learning~\cite{kumar2010selfpaced} strategy by first learning from easy samples and then gradually increase the difficulty of the training samples. We show that, by explicitly modeling the uncertainty on the pose of diverse supervised samples, our approach significantly outperforms both the baselines under various experimental settings.


\section{Related Work}
\label{sec:related_work}
With the introduction of ``DeepPose'' by Toshev \etal~\cite{toshev2014deeppose}, research on human pose estimation began to shift from classic approaches based on pictorial structures \cite{andriluka2009pictorial, bourdev2009poselets, ferrari2008progressive, johnson2011learning, ladicky2013human, pishchulin2013strong, ramanan2006learning, sapp2013modec, yang2013articulated} to deep networks. Subsequent methods include~\cite{tompson2014joint}, which simultaneously captures features at a variety of scales using heatmaps, and~\cite{wei2016convolutional}, which employs a hierarchical model to capture the relationships between joints. A popular approach by Newell \etal~\cite{newell2016stacked} uses conv-deconv architecture and residual model to efficiently generate the heatmap without the need for any hierarchical processing. This approach has been further extended by using visual attention~\cite{Chu_2017_CVPR} and feature pyramid~\cite{yang2017learning}. However, these methods rely on the network capacity to capture the highly articulated human pose and to handle occlusion, without modeling the uncertainty in pose explicitly.

Modeling the uncertainty over the human pose becomes crucial in a diverse data setting, where some of the training samples only provide action information. While pose has often been used to predict action~\cite{Lillo_2016_CVPR, thurau2008pose,vemulapalli2014human,Vemulapalli_2016_CVPR}, the use of action for pose estimation has largely been explored for either 3D human pose~\cite{yao2012coupled}, or for videos where there is temporal information available~\cite{DBLP:journals/corr/IqbalGG16, raja2011joint,xiaohan2015joint, yu2010real}. To the best of our knowledge, our work is the first to exploit action information for 2D pose estimation in still images.

While the specific problem of pose estimation using action information has not been the subject of much attention, the general problem of diverse data learning has a rich history in machine learning and computer vision. Most of the traditional approaches relied on the use of simple parametric structured models such as conditional random fields, or structured support vector machines~\cite{bouchacourt2015entropy, kumar2012modeling, miller2012max, ping2014marginal, schwing2012efficient, yu2009learning}. These methods framed the task of predicting the missing information as estimating latent variables, and employed either the maximum likelihood or the max-margin formulation to efficiently estimate the parameters of the corresponding models. However, as the traditional structured prediction models have now been replaced by deep learning, the aforementioned formulations would need to be adapted for parameter estimation of neural networks. Indeed, our work can be viewed as a natural generalization of~\cite{kumar2012modeling} for deep probabilistic models that admit efficient sampling mechanisms.

The deep learning community also realizes the importance of using diverse data sets to scale-up the data hungry neural network based approaches. This has lead to recent research in deep multiple instance learning~\cite{Durand_2016_CVPR, pathak2014fully, pinheiro2015image}, as well as expectation-maximization based methods~\cite{papandreou2015weakly,pathak2015constrained}. However, most of the deep diverse data learning approaches have been designed to work for a specific task, such as semantic segmentation~\cite{kolesnikov2016seed,tokmakov2016weakly}. It is not clear how the proposed methods can be adapted to learn human poses from action labels. In contrast, our general formulation (presented in the next section) can be easily adapted to any task by simply specifying a task-specific loss function. While we are primarily interested in pose estimation, our formulation may be of interest to the broader audience working on diverse data deep learning.


\section{Problem Formulation}
Our approach uses the recently proposed deep probabilistic network, {\sc Disco} Net~\cite{bouchacourt2016disco}. The {\sc Disco} Net framework allows us to adapt a pointwise network (that is, a network that provides a single pointwise prediction) to a probabilistic one by introducing a noise filter in the pointwise network.

\begin{figure*}[h!]
	\begin{center}
		\includegraphics[scale=0.3]{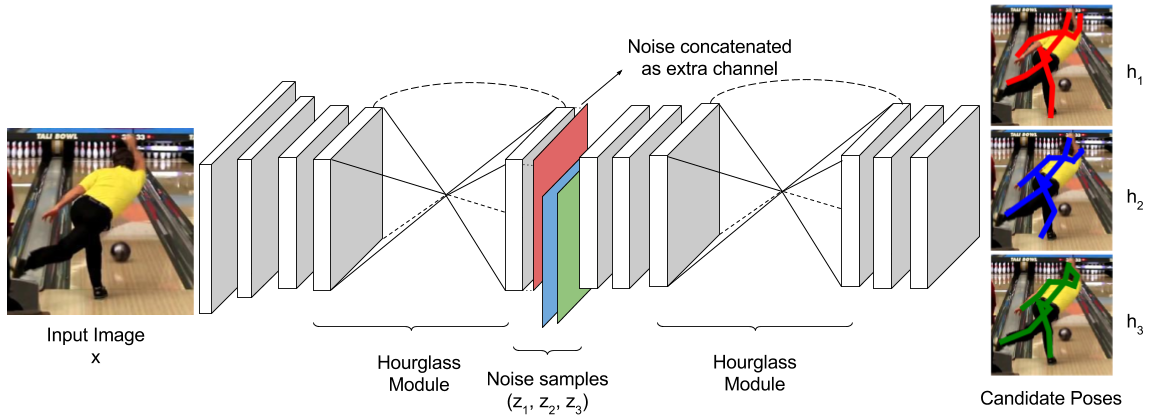}
	\end{center}
    \vspace{-5mm}
   	\caption{{\em For a single input image ${\bf x}$ and three different noise samples $\{ {\bf z}_1, {\bf z}_2, {\bf z}_3\}$ (represented as red, green, blue matrix respectively), \textsc{Disco} Nets produces three different candidate poses $\{ {\bf h}_1, {\bf h}_2, {\bf h}_3\}$. Here each block is a residual layer and two hourglass shaped blocks represent the hourglass module proposed by Newell \etal~\cite{newell2016stacked}. Best viewed in color.}}
    \label{fig:disco-hg}
    \vspace{-3mm}
\end{figure*}

As a concrete example, consider the modified stacked hourglass network in figure~\ref{fig:disco-hg}, which can be used for human pose estimation. The colored filters in the middle of the network represent the noise that is sampled from a uniform distribution. Each value of the noise filter results in a different pose estimate for the same image, thereby enabling us to generate samples from the underlying distribution encoded by the network parameters. Note that obtaining a single sample is as efficient as a forward pass through the network. By placing the filters sufficiently far away from the output layer of the network, we can learn a highly non-linear mapping from the uniform distribution (used to generate the noise filter) to the output distribution (used to generate the pose estimates).

In~\cite{bouchacourt2016disco}, the parameters of a {\sc Disco} Net were learned by minimizing the dissimilarity of the network distribution and the true distribution (as specified by fully supervised training samples). However, we show how the {\sc Disco} Net framework can be extended to enable diverse data learning.


\subsection{Model}
\label{ssec:model}
Due to the uncertainty inherent in the task of pose estimation (occlusion of joints, articulation of human body) as well as the uncertainty introduced by the use of a diverse data set during training, we advocate the use of a probabilistic formulation. To this end, we define two distributions. The first is the prediction distribution that models the probability of a pose ${\bf h}$ given an image ${\bf x}$. As the name suggests, this distribution is used to make a prediction during test time. In this work, we model the prediction distribution $\Pr_{\bf w}({\bf h}|{\bf x})$ as a DISCO Net, where ${\bf w}$ are the parameters of the network.

In addition to the prediction distribution, we also model a conditional distribution of the pose given the image and the action label. As the conditional distribution contains additional information, it can be expected to provide better pose estimates. We will use this property during training to learn an accurate prediction distribution using the conditional distribution. As will be seen shortly, the conditional distribution will not be used during testing. Similar to the prediction distribution, the conditional distribution $\Pr_{\bm{\theta}}({\bf h}|{\bf x},{\bf a})$ is modeled using a DISCO Net, with parameters $\bm{\theta}$. Note that, while we do not have access to the partition function of the two aforementioned distributions, the use of a DISCO Net ensures that we can efficiently sample from them. This property will be exploited to make both the testing and the training computationally feasible.


\subsection{Prediction}
\label{ssec:prediction}
Throughout the rest of the paper, we will assume a task-specific loss function $\Delta(\cdot,\cdot)$ that measures the difference between two putative poses of an image. Given an image ${\bf x}$ containing a human, we would like to estimate the pose ${\bf h}$ of the human such that it minimizes the risk of prediction (as measured by the loss function $\Delta$). Since the ground-truth pose is unknown, we use the principle of maximum expected utility (\textsc{meu})~\cite{premachandran2014empirical}. The {\sc meu} criterion minimizes the expected loss using a set of samples ${\cal H} = \{{\bf h}^k, k=1,\dots,K\}$ obtained from the distribution $\Pr_{\bf w}({\bf h}|{\bf x})$. 

Formally, given an image {\bf x}, we provide a pointwise prediction of the pose in two steps. First, we estimate $K$ pose samples using $K$ different noise filters, each of which is sampled from a uniform distribution. Second, we use the {\sc meu} criterion to obtain the prediction as,
\begin{equation}
	{\bf h}_{\Delta}^*({\bf x};{\bf w}) = \argmin_{k \in [1,K]} \sum_{k'=1}^K \Delta({\bf h}^k, {\bf h}^{k'}).
    \label{eq:h_predict}
\end{equation}
As can be seen, the above criterion can be easily applied for any loss function. For human pose estimation, we adopt the commonly used loss function that measures the mean squared error between the belief maps of two poses over all the joints \cite{newell2016stacked, tompson2014joint, wei2016convolutional}. The belief map $b_{\bf h}(j)$ of a joint $j$ is created by defining a 2D Gaussian whose mean is at the estimated location of the joint, and whose standard deviation is a fixed constant. 


\subsection{Diverse Data Set}
\label{ssec:diverse_data_set}
In order to learn the parameters {\bf w} of the prediction distribution, we require a training data set. Current methods rely on a fully supervised setting, where each training sample is labeled with its ground-truth pose. In order to avoid the cost of such detailed annotations, we advocate the collection of a diverse data set, with a small number of fully supervised samples and a large number of weakly supervised samples. The presence of fully supervised samples helps disambiguate the problem of pose estimation from the problem of action classification.

Formally, we denote our training data set as $\mathcal{D} = \{\mathcal{W}, \mathcal{S}\}$, where $\mathcal{W} = \{({\bf x}_i, {\bf a}_i), i = 1 \dots n\}$ is the weakly annotated data set, and $\mathcal{S}=\{({\bf x}_j, {\bf a}_j, {\bf h}_j), j = 1 \dots m \}$ is the strongly annotated data set and $m < n$.  Here ${\bf x}_i$ refers to the $i$-th training image and ${\bf a}_i$ denotes its action. We denote the underlying pose of the image ${\bf x}_i$ as the latent variable ${\bf h}_i$. Note that we do not assume a single underlying pose. Instead, we model the distribution over all putative poses given the image and the action.


\subsection{Learning Objective}
\label{ssec:learning_objective}
Given the diverse data set $\mathcal{D}$, our goal is to learn the parameters ${\bf w}$ such that it provides an accurate pose estimate ${\bf h}^{*}_{\Delta}({\bf x};{\bf w})$ (specified in equation~(\ref{eq:h_predict})) for a test image {\bf x}. A typical learning objective for this purpose would estimate the joint distribution $\Pr_{\bf w}({\bf h},{\bf a}|{\bf x})$ using expectation-maximization or its variants~\cite{bishop2007pattern}. Given an image {\bf x}, the pose would then be obtained by marginalizing over all actions {\bf a}. However, we argue that this approach needlessly places the burden of accurately representing the uncertainty of the pose and the action of an image on a single distribution. Since the action information would not be provided during testing, such an approach may fail to fully utilize the modeling capacity of the distribution parameters to obtain the best pose.

Inspired by the work of Kumar \etal~\cite{kumar2012modeling}, we design a joint learning objective that minimizes the dissimilarity coefficient between the prediction distribution and the conditional distribution. Briefly, the dissimilarity coefficient between two distributions $\Pr_1(\cdot)$ and $\Pr_2(\cdot)$ is determined by measuring their diversities. The diversity coefficient of a distribution $\Pr_1(\cdot)$ and a distribution $\Pr_2(\cdot)$ is defined as the expected difference between their samples, where the difference is measured by any task-specific loss function $\Delta'(\cdot,\cdot)$. Formally, we define the diversity coefficient as,
\begin{equation}
	\text{DIV}_{\Delta'}(\Pr\nolimits_1,\Pr\nolimits_2) = \sum_{{\bf y}_1,{\bf y}_2 \in {\cal Y}}\Delta'({\bf y}_1,{\bf y}_2)\Pr\nolimits_1({\bf y}_1)\Pr\nolimits_2({\bf y}_2).
	\label{eq:div}
\end{equation}
where ${\cal Y}$ is the space over which the distributions are defined.
Using the definition of diversity, the dissimilarity coefficient of $\Pr_1$ and $\Pr_2$ is given by,
\begin{equation}
	\text{DISC}_{\Delta'}(\Pr\nolimits_1,\Pr\nolimits_2) = \text{DIV}_{\Delta'}(\Pr\nolimits_1,\Pr\nolimits_2) - \gamma\text{DIV}_{\Delta'}(\Pr\nolimits_1,\Pr\nolimits_1) - (1-\gamma)\text{DIV}_{\Delta'}(\Pr\nolimits_2,\Pr\nolimits_2).
    \label{eq:disco}
\end{equation}
In other words, the dissimilarity between $\Pr\nolimits_1$ and $\Pr\nolimits_2$ is the difference between the diversity of $\Pr_1$ and $\Pr\nolimits_2$ and an affine combination of their self-diversities. In our experiments, we use $\gamma = 0.5$, which results in a symmetric dissimilarity coefficient between two distributions.

Given the above definition, we can now specify our learning objective as,
\begin{eqnarray}
	\argmin_{{\bf w},\bm{\theta}} \sum_{i=1}^n DISC_{\Delta}(\Pr\nolimits_{\bf w}(\cdot|{\bf x}_i),\Pr\nolimits_{\bm{\theta}}(\cdot|{\bf x}_i,{\bf a}_i)).
	\label{eq:learn_obj}
\end{eqnarray}
In other words, our learning objective encourages the prediction distribution and the conditional distribution to agree with each other (that is, have a small dissimilarity coefficient) for all training samples. Intuitively, the conditional distribution $\Pr_{\bm{\theta}}(\cdot|{\bf x},{\bf a})$ would be able to significantly narrow down the set of probable poses for a given image using the action information. By minimizing the dissimilarity between the prediction distribution and the conditional distribution, our learning objective will encourage the prediction to assign a high probability to the set of poses that are compatible with the given action. During testing, only the prediction distribution will be used to obtain the pose of a given image.

Computationally, the main challenge of employing the learning objective~(\ref{eq:learn_obj}) is that its value can only be determined by estimating the loss function over all possible pairs of poses. However, the key observation that enables its use in practice is that we can obtain an unbiased estimate of its value, as well as its gradient, by sampling from the distributions $\Pr_{\bf w}$ and $\Pr_{\bm{\theta}}$. In other words, given samples $\{{\bf h}_k,k=1,\cdots,K\}$ from the prediction distribution, and samples $\{{\bf h}'_k,k=1,\cdots,K\}$ from the conditional distribution, the unbiased estimated value of the learning objective~(\ref{eq:learn_obj}) can be computed as,
\begin{eqnarray}
	\frac{1}{K^2}\left(\sum_{k,k'}\Delta({\bf h}_k,{\bf h}'_{k'})
-\gamma\sum_{k,k'}\Delta({\bf h}_k,{\bf h}_{k'}) -(1-\gamma)\sum_{k,k'}\Delta({\bf h}'_k,{\bf h}'_{k'})
\right).
	\label{eq:unbiasedObj}
\end{eqnarray}


\subsection{Optimization}
\label{ssec:optimization}
\begin{figure}[!ht]
	\begin{center}
		\includegraphics[scale=0.25]{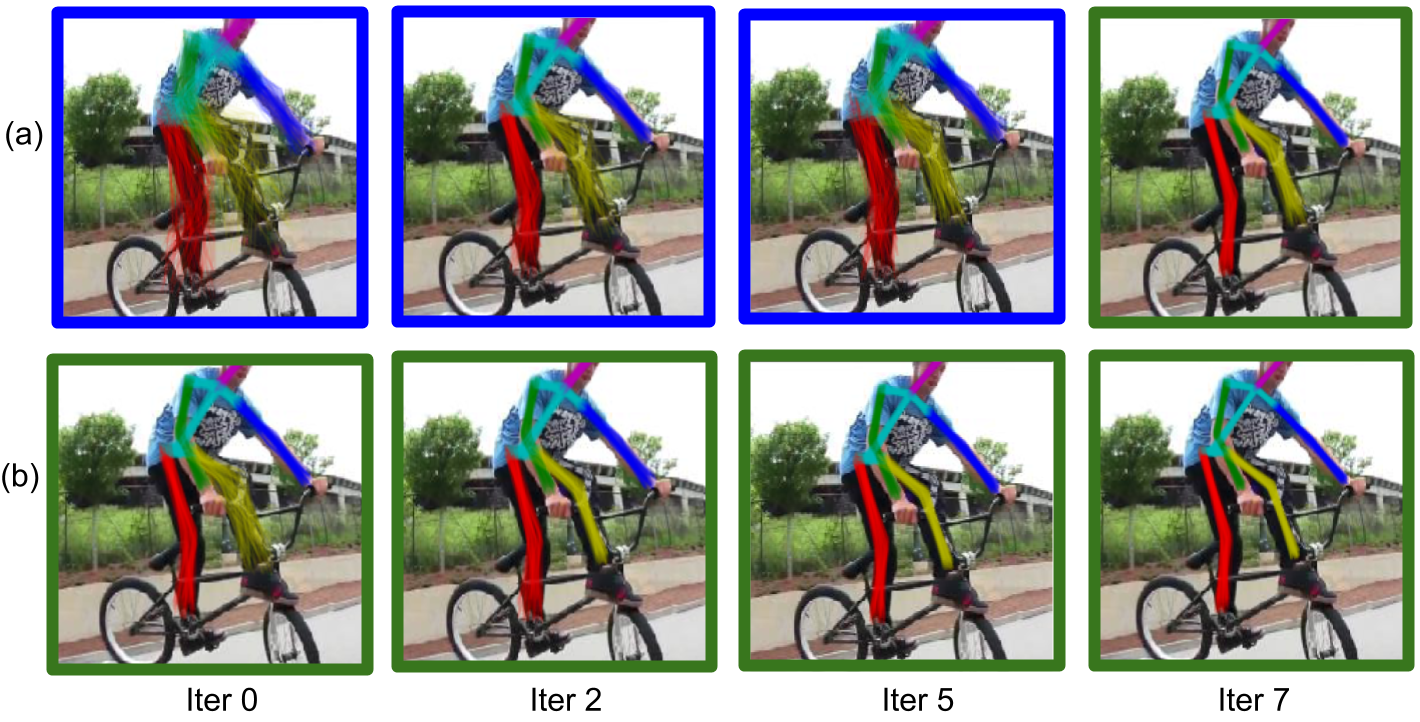}
	\end{center}
    \vspace{-5mm}
   	\caption{\em{Example of superimposed pose predictions by \textsc{Disco} Nets illustrating the uncertainty in the pose across training iterations. The blue box around the images represent a high diversity coefficient value, and the green box around them represents low diversity coefficient value. Row (a) represents outputs from the prediction network and row (b) represents outputs from the conditional network. The first column shows the initial prediction of the networks; columns $2$ through $4$ shows prediction of the networks at second, fifth and final iteration respectively. The images show a common action of riding a bike where the conditional network performs well from the beginning of the optimization procedure and transfers its knowledge to the prediction network. Best viewed in color.}}
    \label{fig:visualization-of-prediction}
    \vspace{-3mm}
\end{figure}

As a {\sc Disco} Net provides an efficient sampling mechanism, it is ideally suited to stochastic gradient descent. In order to make the most use of the diverse nature of the data set, as well as the learning objective, we estimate the parameters of the two networks in three stages. First, we use supervised training for the two networks using the small amount of the ground truth pose data. Second, we perform iterative training of the two networks, that is, we update one network while keeping the other fixed. Third, we jointly optimization of both the networks together. At each stage, we use stochastic gradient descent in a similar manner to~\cite{bouchacourt2016disco}. Joint training of the two network is expensive in terms of memory and time. However, by first training the two networks using strong supervision and then using iterative optimization strategy, we significantly reduce the number of iterations required in the third stage of the optimization. We provide further details in the supplementary.

The prediction of the two networks during the iterative training stage is visualized in figure~\ref{fig:visualization-of-prediction}. For a commonly occurring action of riding a bike, we depict the hundred different pose estimates from the prediction and the conditional network by superimposing them. Hence, if all the pose estimates agree with each other, the lines depicting the samples will be thin and opaque. In order to represent the low uncertainty in the pose estimates of this image, we will draw a green bounding box around the image. In contrast, if the pose estimates vary significantly from each other, then the lines depicting the samples will be spread out and less opaque. In order to represent the high uncertainty in the pose estimates of this image, we will draw a blue bounding box around the image. 

Here, we observe that initially $\Pr_{\bf w}$ has high uncertainty for the predicted pose, but $\Pr_{\bm{\theta}}$ is confident about the predictions. However, after several iterations of the optimization algorithm, the information present in the conditional network is successfully transferred over to the prediction network. This is shown in the last column, where both the networks start to agree with each other (that is, have a low self diversity coefficient). For difficult images where both prediction and conditional distribution are highly uncertain at the beginning, the networks learns from other easy examples that may be present in the data set. Further visualizations of the learning process are provided in the supplementary material.

\section{Experiments}
\label{sec:experiments}
\paragraph{Data set.} We use the {\sc MPII} Human pose data set \cite{andriluka20142d}, which consists of $17.4$k images with publicly available action and ground-truth pose annotations. We split the images into $\{70,15,15\}\%$ training, validation and test sets, which corresponds to $12,156$ images in the training set and $2605$ images each in the testing and the validation set. In order to obtain a diverse data set, we discard the pose information for a random subset of training examples, while retaining action labels for all samples. This results in (i) a fully annotated training set, which contains both the ground truth pose annotations and the action labels; and (ii) a weakly annotated training set, which only contains action labels. 

To obtain the tasks of varying levels of difficulty, we choose three different data splits, $\{25-75, 50-50, 75-25\}\%$, where we randomly discard $75\%$, $50\%$, and $25\%$ of the pose annotations from the training images respectively. We note here that for each split, we augment our training set by rotating the images with an angle ($+/- 30^{\circ}$) and by horizontal flipping the original image. 


\paragraph{Implementation Details.} In order to implement our probabilistic {\sc Disco} network, shown in figure~\ref{fig:disco-hg}, we adopt the popular stacked hourglass network~\cite{newell2016stacked} for human pose estimation, which stacks $8$ hourglass modules. For the prediction network, a noise filter of size $64 \times 64$ is added to the output of the penultimate hourglass module, which itself consists of $256$ $64 \times 64$ filters. The $257$ channels are convolved with a $1 \times 1$ filter to bring the number of channels back to $256$. This is followed by a final hourglass module as shown in figure~\ref{fig:disco-hg} (closely following the approach of~\cite{newell2016stacked}). As noise is treated as input, all parameters of the network remain differentiable and hence can be trained via backpropagation. Our conditional network is modeled exactly as the prediction network, except that there are ${\bf a}$ different output branches, one for each possible action class, stacked on top of penultimate hourglass module. Each output branch has its own noise filter followed by the final hourglass module as described before. We present additional implementation details in the supplementary.

Notice that when drawing $K$ samples from this modified stacked hourglass architecture for the same input image, we can reuse the output of the penultimate layer of the 8-stacked hourglass net. We only need to recompute the final hourglass module $K$ times to generate $K$ samples which greatly reduces our runtime complexity. In practice, a single forward pass to draw $K = 100$ samples from our probabilistic net takes 114 ms. compared to 68 ms. for the vanilla stacked hourglass network on NVIDIA GTX 1080Ti GPU.


Initialization of the prediction network is done by training it using a small number of fully annotated training data, while for the conditional networks, we initialize them by fine tuning the prediction network weights with the small number of action specific fully annotated training data. We then optimize our two set of networks by first, iterative optimization procedure, and then through joint optimization, as described in the previous section.


\paragraph{Methods.} We compare our proposed probabilistic method, learned with diverse data, with two baselines: (i) a fully supervised human pose estimation network, the stacked hourglass network~\cite{newell2016stacked}, which we refer to as {\sc FS} Net; and (ii) a non-probabilistic pointwise network trained with diverse data, which uses the same architecture as shown in figure~\ref{fig:disco-hg} but provides a single prediction.  We refer this pointwise network as {\sc PW} Net. The first baseline helps us to compare the performance of a fully supervised network with a network trained on the diverse collection of data, and the second baseline demonstrates the benefit of our probabilistic network when compared to a non probabilistic pointwise network.


We train {\sc FS} net on the fully annotated data set using stochastic gradient descent, as discussed in~\cite{newell2016stacked}. The {\sc PW} net is trained using diverse data, making use of the action annotations. We provide the detailed training setup of the {\sc FS} and the {\sc PW} net in the supplementary.

\begin{table}[t]
\centering
\resizebox{0.9\textwidth}{!}{%
\begin{tabular}{l|l|lllllll|l}
\hline
Method & Split & Head & Sho. & Elb. & Wri. & Hip & Knee & Ank. & Total \\ 
\hline
Supervised Subset	 & 100\% & 98.16 & 96.22 & 91.23 & 87.08 & 90.11 & 87.39 & 83.55 & 90.92 \\
\hline
\multirow{3}{*}{FS}  & 25\% & 59.17 & 46.98 & 30.00 & 21.33 & 36.32 & 20.05 & 23.93 & 37.54 \\
                     & 50\% & 90.18 & 80.60 & 64.29 & 52.43 & 67.44 & 55.41 & 51.30 & 67.88 \\
                     & 75\% & 94.61 & 90.56 & 81.28 & 74.15 & 81.86 & 73.20 & 67.19 & 80.88 \\
\hline 
\multirow{3}{*}{PW} & 25-75 & 73.77 & 55.69 & 37.21 & 25.32 & 43.24 & 28.01 & 30.82 & 45.16 \\
                    & 50-50 & 92.97 & 83.56 & 71.08 & 59.18 & 72.56 & 60.49 & 57.27 & 73.11 \\
                    & 75-25 & 95.46 & 93.50 & 86.47 & 81.05 & 85.58 & 80.98 & 76.81 & 85.89 \\
\hline 
\multirow{3}{*}{$\Pr\nolimits_{\bf w}$ (iterative)} 
					& 25-75 & 78.21 & 60.98 & 42.01 & 28.75 & 42.37 & 29.07 & 33.54 & 48.12 \\
                    & 50-50 & 93.42 & 86.91 & 75.03 & 66.56 & 77.22 & 67.38 & 60.96 & 76.43 \\
                    & 75-25 & 96.28 & 94.53 & 88.36 & 83.31 & 87.54 & 82.45 & 79.48 & 88.16 \\       
\hline 
\multirow{3}{*}{$\Pr\nolimits_{\bf w}$ (joint)} 
			& 25-75 & 79.54 & 62.87 & 43.38 & 29.38 & 43.38 & 30.91 & 34.86 & \textbf{49.41} \\
            & 50-50 & 94.07 & 88.32 & 75.93 & 67.53 & 78.20 & 67.80 & 61.49 & \textbf{78.01} \\
            & 75-25 & 97.45 & 95.87 & 90.21 & 86.09 & 89.42 & 86.26 & 82.92 & \textbf{90.21} \\
\hline 
\end{tabular}
}
\vspace{2mm}
\caption{{\em Results on \textsc{MPII} Human Pose ({\em \textsc{PCK}h@0.5}), where {\sc FS} is trained on varying percentages of fully annotated data and {\sc PW} and {\em $\Pr\nolimits_{\bf w}$} are trained on varying splits of fully annotated and weakly annotated training data. Here {\sc FS} and {\sc PW} are the fully supervised and the pointwise networks respectively, and {\em $\Pr\nolimits_{\bf w}$} (iterative) and {\em $\Pr\nolimits_{\bf w}$} (joint) is our proposed probabilistic network trained with iterative optimization and joint optimization respectively. The supervised subset is the fully supervised stacked hourglass net~\cite{newell2016stacked} trained with all the available labels and defines the upper bound on the total accuracy that can be achieved through this architecture.}}
\label{table:results}
\vspace{-3mm}
\end{table}


\paragraph{Results.} We evaluate the three trained networks, {\sc FS}, {\sc PW} and $\Pr\nolimits_{\bf w}$, by computing their accuracy on the held out test set. We use the normalized ``Probability of Correct Keypoint'' ({\sc PCK}h) metric~\cite{sapp2013modec} to report our results. Table \ref{table:results} shows the performance of the three networks when trained on varying splits of the training set.

Here, we observe that, for all the data splits, our proposed probabilistic network $\text{Prob}_{\bf w}$ outperforms the other baseline networks {\sc FS} and {\sc PW}. This superior performance is seen consistently across predictions of all joints as well as on the overall pose prediction.

Performance of the three networks, {\sc FS}, {\sc PW} and $\Pr\nolimits_{\bf w}$, increases with the increase in level of supervision. In the more challenging $25-75$ split, there are far fewer fully supervised examples present for each action category which results in {\sc PW} and $\Pr\nolimits_{\bf w}$ to learn a poor initial estimate of action specific pose from diverse data. This leads to overall poor performance when compared to $50-50$ or $75-25$ split case, where we have more supervised data.

Moreover, both the methods trained using diverse data, {\sc PW} and $\Pr\nolimits_{\bf w}$, show a significant gain in accuracies when compared to the fully supervised network, {\sc FS}. This empirically shows us that the action information present in the weakly annotated set is helpful for predicting pose. 

As our proposed probabilistic network $\Pr\nolimits_{\bf w}$ performs better than the pointwise network {\sc PW}, we see the significance of modeling uncertainty over pose. Though the proposed probabilistic network only marginally improves the prediction for joints with low uncertainty, like the head, shoulder and hips, the difference in the accuracies of the two networks is due to better performance of the probabilistic network $\Pr\nolimits_{\bf w}$ on difficult joints like wrists, elbows, knees and ankles. We see that the $\Pr\nolimits_{\bf w}$ network provides a significant improvement of up to $5\%$ improvement in accuracies over the {\sc PW} Net on joints with high uncertainty (wrists, elbows, ankles and knees). 

Joint training of the two set of networks improves our prediction by around $1.5\%$. We also note that, while the supervised subset, which is the fully supervised stacked hourglass network~\cite{newell2016stacked} trained using all available labels in the training set, achieves $90.9\%$ \cite{newell2016stacked}, our probabilistic network provides comparable results when trained only on $75\%$ pose annotations and $25\%$ action annotations. Note that the supervised subset defines the upper bound on accuracy that can be achieved through this architecture.

We argue that the relative position of joints like head, shoulder and hip remains largely in similar spatial location with respect to each other across various actions and therefore have low entropy, whereas, joints like wrists, elbows, knees and ankles not only show huge variations in their relative spatial location across various action categories but also within same action category, resulting in large entropy. Therefore, even though pointwise network {\sc PW} does a good job of estimating pose locations for joints with low uncertainty, it fails to capture the high inter-class and intra-class variability of joints with high uncertainty. On the other hand,  $\Pr\nolimits_{\bf w}$ explicitly models uncertainty over joint locations as can be seen in figure~\ref{fig:motivation}.

Our method was implemented using PyTorch library\footnote{The code and the pre-trained model is available at \url{http://bit.ly/poses-from-actions}}. Further details of the experimental setup, full PcKh curves, and results for additional experiments using a different architecture~\cite{belagiannis2015robust}, demonstrating the generality of our method, are included in the supplementary material.

\vspace{-3mm}
\section{Discussion}
\label{sec:discussion}
We presented a novel framework to learn human pose using diverse data set. Our framework uses two separate distributions: (i) a conditional distribution for modeling uncertainty over pose given the image and the action during training time; and (ii) a prediction distribution to provide pose estimates for a given image. We model the two aforementioned distributions using a deep probabilistic network. We learn these separate yet complimentary distributions by minimizing a dissimilarity coefficient based learning objective. Empirically, we show that: (i) action serves as an important cue for predicting human pose; and (ii) modeling uncertainty over pose is essential for its accurate prediction. 

Our approach can be easily adapted to other diverse learning tasks by specifying an appropriate loss function for the evaluation of the diversity coefficient. This may be of interest to a wider machine learning and computer vision audience. We would also like to investigate the the use of active learning, so that our network benefits the most in terms of accuracy from the fully supervised annotations. The diversity of the pose samples, which can be computed efficiently in our framework, can provide a useful cue to enable active learning.

\vspace{-3mm}
\section{Acknowledgements}
\label{sec:acknowledgements}
This work is partially funded by the EPSRC grants EP/P020658/1 and TU/B/000048 and a CEFIPRA grant. Aditya is supported by Visvesvaraya Ph.D. Fellowship program. 

\clearpage
\bibliography{egbib}
\clearpage

\part*{Supplementary Material}

\appendix

\section{Optimization}
\label{supsec:optimization}
In this section, we provide details of optimization presented in section~\ref{ssec:optimization} above. 

\subsection{Learning Objective}
We represent the prediction distribution using a {\sc Disco} Net, which we denote by $\Pr\nolimits_{\bf w}$, {\bf w} being the parameter of the network. Similarly, we represent the conditional distribution using a set of {\sc Disco} Nets, which we denote by $\Pr\nolimits_{\boldsymbol{\theta}}$. The set of parameters for the conditional networks is denoted by $\boldsymbol{\theta}$. We compute samples from the prediction network as $\{{\bf h}^{\bf w}_k,k=1,\cdots,K\}$, and samples from conditional network as $\{{\bf h}'^{\bf \boldsymbol{\theta}}_{k},k=1,\cdots,K\}$ for a given training sample. The unbiased estimated value of the learning objective (\ref{eq:unbiasedObj}) can be written as follows:
\begin{eqnarray}
\argmin_{{\bf w},\mathbf{\theta}}F({\bf w}, \boldsymbol{\theta}) =  \frac{1}{NK^2}\sum_{i=1}^N\left(\sum_{k,k'}\Delta({\bf h}_k^{\bf w},{\bf h}_{k'}^{\bf \theta}) -\gamma\sum_{k,k'}\Delta({\bf h}_k^{\bf w},{\bf h}_{k'}^{\bf w}) \right. \nonumber \\
\left. -(1-\gamma)\sum_{k,k'}\Delta({\bf h}_k^{\bf \theta},{\bf h}_{k'}^{\bf \theta}) \right)	
\label{eq:joint_objective}
\end{eqnarray}

In order to minimize the dissimilarity coefficient between the parameters of the prediction and the conditional distributions, we employ stochastic gradient descent. We note that jointly optimizing the objective function over the parameters of the prediction and the conditional distribution networks is expensive in terms of memory and time, as it involves optimizing two networks together. Therefore, first, we initialize the two networks by training them with the small amount of fully annotated pose data. We then perform iterative optimization using block coordinate descent to first train the parameters of the prediction and conditional distribution and then proceed with more expensive joint optimization. Algorithm for optimizing these two sets of parameters are shown in the following subsections. Using this hybrid training strategy, we reduce the training complexity without compromising on the accuracy.

\subsection{Iterative Optimization}
\label{supssec:iterative_optimization}
The coordinate descent optimization proceeds by iteratively fixing the prediction network and estimating the conditional networks, followed by updating the prediction network for fixed conditional networks. The parameters of both the set of networks are initialized using the small amount of fully supervised samples available in the data set. The main advantage of the iterative strategy is that it results in a problem similar to the fully supervised learning of {\sc Disco} Nets at each iteration. This, in turn, allows us to readily use the algorithm developed in~\cite{bouchacourt2016disco}. Furthermore, it also reduces the memory complexity of learning, thereby allowing us to learn a large network. The two steps of the iterative algorithm are described below. 

\paragraph{Optimization over Conditional Network}
For fixed ${\bf w}$, the learning objective corresponds to the following:
\begin{equation}
\argmin_{\bm{\theta}} \sum_i DIV(\Pr_{\bf w},\Pr_{\bm{\theta}}) - 
(1-\gamma)DIV(\Pr_{\bm{\theta}},\Pr_{\bm{\theta}})
\label{eq:theta_objective_fn}
\end{equation}
The above equation can be expanded as,
\begin{eqnarray}
\min_{\boldsymbol{\theta}}F({\boldsymbol{\theta}}) = \frac{1}{NK^2} \sum_{i=1}^N\left(\sum_{k,k'}\Delta({\bf h}_k^{\bf w},{\bf h}_{k'}^{\bf \theta}) -(1-\gamma)\sum_{k,k'}\Delta({\bf h}_k^{\boldsymbol{\theta}},{\bf h}_{k'}^{\boldsymbol{\theta}}) \right)
\label{eq:theta_objective_fn}
\end{eqnarray}
The above objective function is similar to the one used in~\cite{bouchacourt2016disco} for fully supervised learning. Similar to~\cite{bouchacourt2016disco}, we solve it via stochastic gradient descent. Note that since it is possible to generate samples from both the prediction and the conditional network, we can obtain an unbiased estimate of the gradient of the objective function~(\ref{eq:theta_objective_fn}). As observed in~\cite{bouchacourt2016disco}, this is sufficient to minimize the learning objective in order to estimate the {\sc Disco} Net parameters.

The above objective function is solved via stochastic gradient descent, as shown in Algorithm~\ref{alg:algo_theta}.
\begin{algorithm}
    \caption{Optimization over $\mathbf{\theta}$ }
    \begin{algorithmic}[t]
        \Require Data set $\mathcal{D}$ and initial estimate $\mathbf{\theta}^0$
        \For{$t = 1 \dots T$ \emph{epochs}}
            \State Sample mini-batch of $b$ training example pairs
            \For{$n = 1 \dots b$}
                \State Sample $K$ random noise vectors $\mathbf{z}_k$
                \State Generate $K$ candidate output from $\Pr\nolimits_{\mathbf{w}}(\mathbf{x}, \mathbf{z}_k)$ and $\Pr\nolimits_{\boldsymbol{\theta}}({\bf x}, {\bf z}_k)$
            \EndFor
            \State Compute $F({\bf \theta})$ as given here in equation (\ref{eq:theta_objective_fn}) here.
            \State Update parameters ${\bf \theta}$ via SGD with momentum 
        \EndFor
    \end{algorithmic}
    \label{alg:algo_theta}
\end{algorithm}

\paragraph{Optimization over Prediction Network}
For fixed $\bm{\theta}$, the learning objective corresponds to the following:
\begin{equation}
\min_{{\bf w}} \sum_i DIV(\Pr_{\bf w},\Pr_{\bm{\theta}}) - 
\gamma DIV(\Pr_{\bf w},\Pr_{\bf w})
\label{eq:w_objective_fn}
\end{equation}
The above equation can be expanded as,
\begin{eqnarray}
\min_{\bf w}F({\bf w}) = \frac{1}{NK^2}\sum_{i=1}^N\left(\sum_{k,k'}\Delta({\bf h}_k^{\bf w},{\bf h}_{k'}^{\boldsymbol{\theta}})
- \gamma\sum_{k,k'}\Delta({\bf h}_k^{\bf w},{\bf h}_{k'}^{\bf w})  \right)
\label{eq:w_objective_fn}
\end{eqnarray}
Once again, using the fact that it is possible to obtain unbiased estimates of the gradients of the above objective function, we employ stochastic gradient descent to update the parameters of the prediction network.

Similar to the conditional network, the above objective function is optimized by using stochastic gradient descent as shown in Algorithm~\ref{alg:algo_w}.

\begin{algorithm}
    \caption{Optimization over $\mathbf{w}$ }

    \begin{algorithmic}[h]
        \Require Data set $\mathcal{D}$ and initial estimate $\mathbf{w}^0$
        \For{$t = 1 \dots T$ \emph{epochs}}
            \State Sample mini-batch of $b$ training example pairs
            \For{$n = 1 \dots b$}
                \State Sample $K$ random noise vectors $\mathbf{z}_k$
                \State Generate $K$ candidate output from  $\Pr\nolimits_{\boldsymbol{\theta}}({\bf x}, {\bf z}_k)$ and $\Pr\nolimits_{\bf w}({\bf x}, {\bf z}_k)$.
            \EndFor
            \State Compute $F(\mathbf{w})$ as given in equation (\ref{eq:w_objective_fn}) here.
            \State Update parameters ${\bf w}$ via SGD with momentum 
        \EndFor
    \end{algorithmic}
    \label{alg:algo_w}
\end{algorithm}

\subsection{Joint Optimization}
\label{supssec:joint_optimization}
Although the iterative optimization provides for faster convergence of our objective function, this approach of finding a local minima along one coordinate direction at the current point, in each iteration, often leads to an approximate solution with respect to the optimization problem at hand. To address this problem and find accurate local minima of our non-convex objective (\ref{eq:unbiasedObj}), we perform joint optimization of our objective function by employing stochastic gradient descent to update the parameters of both conditional and prediction distribution networks. We obtain the gradients by computing the unbiased estimate of our objective function and update the two networks using stochastic gradient descent as shown in Algorithm~\ref{alg:algo_joint}. Additionally, we initialize our parameters of the networks corresponding to the two distributions with the values obtained after the iterative optimization. This initialization strategy also reduces the number of iterations required for convergence, thus reducing the training time complexity. 

\begin{algorithm}
    \caption{Joint Optimization over $\mathbf{w}, \boldsymbol{\theta}$ }

    \begin{algorithmic}[h!]
        \Require Data set $\mathcal{D}$, learning rate $\eta$, momentum $m$,\\
        and initial estimate $\mathbf{w}^0, \boldsymbol{\theta}^0$
        \For{$t = 1 \dots T$ \emph{epochs}}
            \State Sample mini-batch of $b$ training example pairs
            \For{$n = 1 \dots b$}
                \State Sample $K$ random noise vectors $\mathbf{z}_k$
                \State Generate $K$ candidate output from  $\Pr\nolimits_{\boldsymbol{\theta}}({\bf x}, {\bf z}_k)$ and $\Pr\nolimits_{\bf w}({\bf x}, {\bf z}_k)$.
            \EndFor
            \State Compute $F(\mathbf{w}, \boldsymbol{\theta})$ as given in equation (\ref{eq:joint_objective}) here.
            \State Update parameters ${\bf w}$ via SGD with momentum 
        \EndFor
    \end{algorithmic}
    \label{alg:algo_joint}
\end{algorithm}

\clearpage

\section{Visualization of the Learning Process}
\label{supsec:visualization_of_the_learning_process}
We provide visualization of the iterative learning procedure as discussed in the optimization section \ref{ssec:optimization}. We show a hundred different pose estimates of two examples, of varying difficulty, over the iterations of the optimization algorithm. The pose estimates are superimposed on the image. Hence, if all the pose estimates agree with each other, the lines depicting the samples will be thin and opaque. In order to represent the low uncertainty in the pose estimates of this image, we will draw a green bounding box around the image. For such images, the expected loss is less than $3$. In contrast, if the pose estimates vary significantly from each other, then the lines depicting the samples will be spread out and less opaque. In order to represent the high uncertainty in the pose estimates of this image, we will draw a blue bounding box around the image. For these samples, the expected loss is more than $3$.

\begin{figure}[!h]
	\begin{center}
		\includegraphics[scale=0.41]{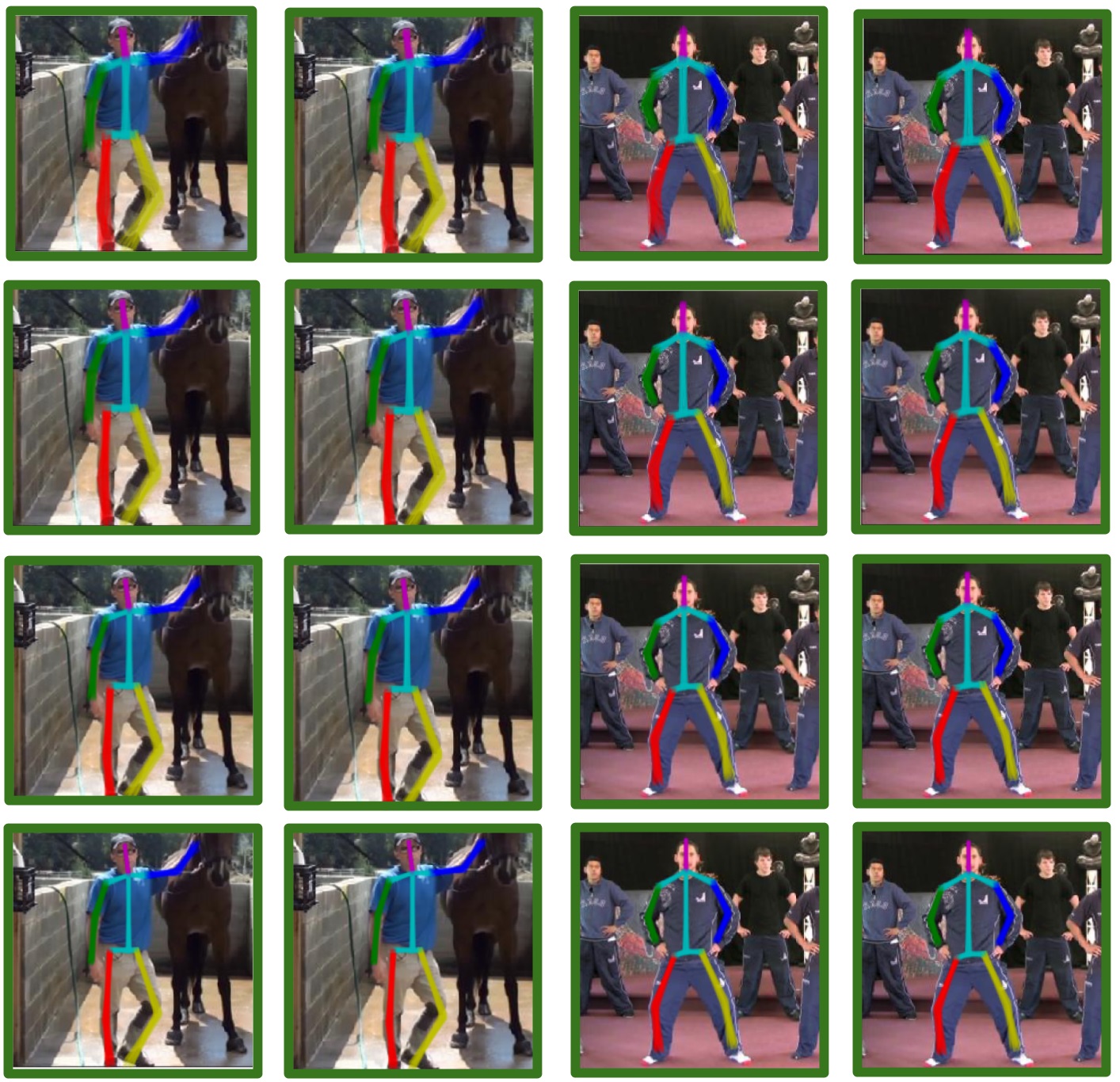}
	\end{center}
   	\caption{\em{Example of superimposed pose predictions by \textsc{Disco} Nets illustrating the uncertainty in the pose across training iterations for an easy case. The blue box around the images represent a high diversity coefficient value, and the green box around them represents low diversity coefficient value. Columns $1$ and $3$ are outputs of the prediction network and columns $2$ and $4$ are outputs of conditional network. Row $1$ represents initial prediction of networks; rows 2 and 3 represents prediction of networks in second and fifth iteration respectively and last row represents prediction of networks when they have converged. The images in the first and second column show an easy example of a person standing straight with his one hand held out and the third and fourth columns show a person standing in relaxed upright pose. where both the conditional network and the prediction network performs well from the beginning of the optimization procedure. For each example, the first column shows estimated pose from prediction network and the second column shows estimated pose from conditional network. Best viewed in color.}}
    \label{fig:visualization-easy}
\end{figure}

The first case shown in figure \ref{fig:visualization-easy} represents an easy case where the initial prediction and conditional networks, $\Pr\nolimits_{\bf w}$ and $\Pr\nolimits_{\boldsymbol{\theta}}$ , trained only on the fully annotated training set, have low uncertainty for the predicted pose. In these images, there are no occlusions of any human part, and the person present in the image is in the standard pose for the particular action he is performing. For such cases, the fully annotated training data set is enough to train the prediction network such that it has high confidence in the estimated pose, and they do not require weakly supervised training. However, even in such cases, we see a minor improvement in the estimated pose over the iterations of the optimization algorithm.

\begin{figure}[!h]
	\begin{center}
		\includegraphics[scale=0.41]{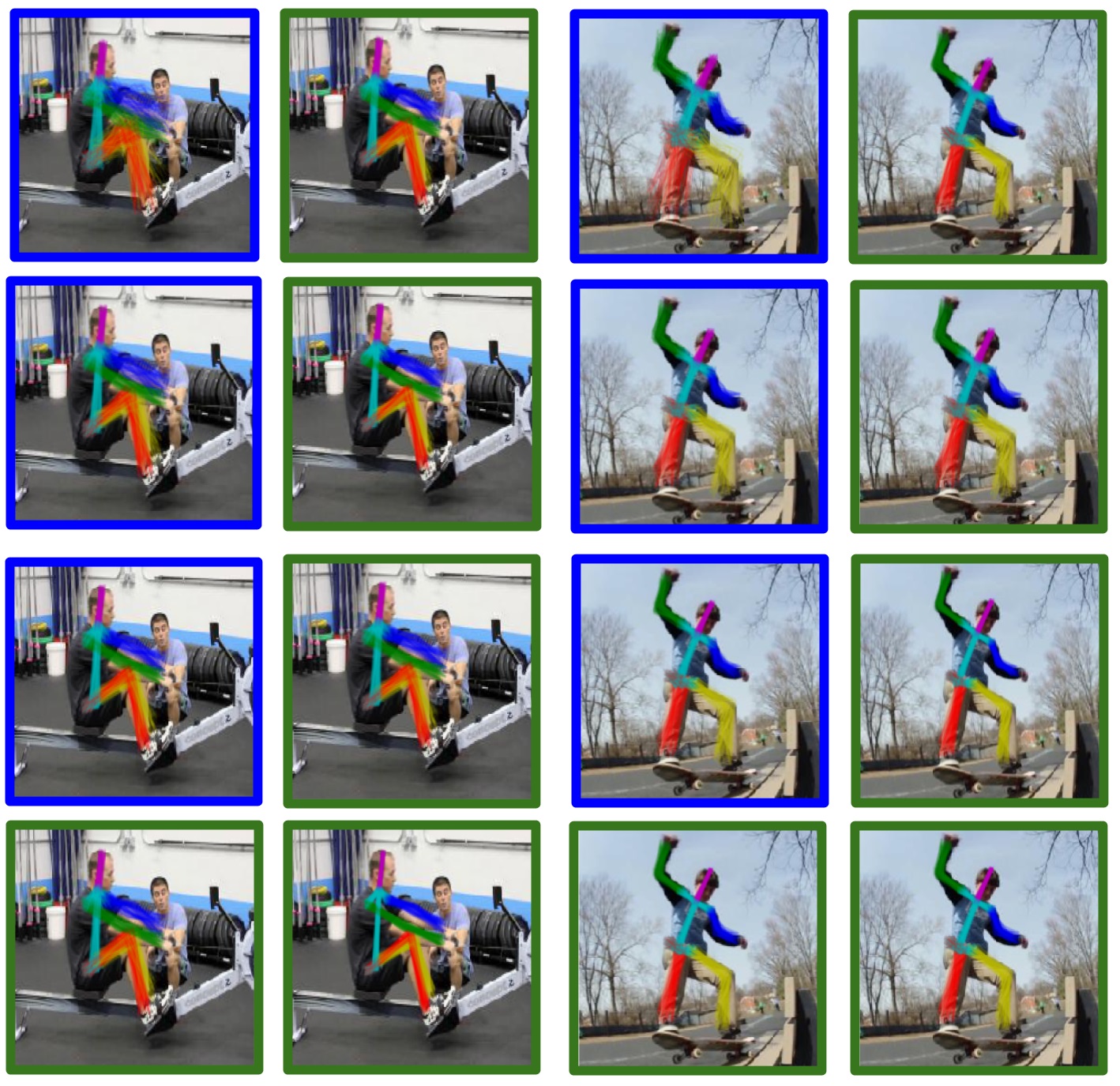}
	\end{center}
   	\caption{\em{Example of superimposed pose predictions by \textsc{Disco} Nets illustrating the uncertainty in the pose across training iterations for examples with moderate difficulty. The blue box around the images represent a high diversity coefficient value, and the green box around them represents low diversity coefficient value. Columns $1$ and $3$ are outputs of the prediction network and columns $2$ and $4$ are outputs of conditional network. Row $1$ represents initial prediction of networks; rows 2 and 3 represents prediction of networks in second and fifth iteration respectively and last row represents prediction of networks when they have converged. The images in the first and second column show a common action of a person exercising and the third and fourth column shows a person riding a skate board. In these cases, the conditional network performs well from the beginning of the optimization procedure. At convergence, both the prediction network provides accurate pose estimates for such moderately difficult images by transferring information from conditional network. For each example, the first column shows estimated pose from prediction network and the second column shows estimated pose from conditional network. Best viewed in color.}}
    \label{fig:visualization-moderate}
\end{figure}

Figure \ref{fig:visualization-moderate} represents a moderately difficult example. Typically, such examples are those where a person is performing commonly occurring actions, like exercising, riding a bike or skate board, or running. In such examples, some joints are occluded and the person in the image is in some variation of the standard pose for a particular action he is performing. The majority of the data set are comprised of moderately difficult examples. In such cases, the prediction network $\Pr\nolimits_{\bf w}$ has high uncertainty over the predicted pose, but conditional network $\Pr\nolimits_{\boldsymbol{\theta}}$ has high confidence and therefore low uncertainty over the predicted pose. Here we observe that over the iterations, the prediction network gains confidence as the information present in the conditional network is successfully transferred to it. 

\begin{figure}[!h]
	\begin{center}
		\includegraphics[scale=0.41]{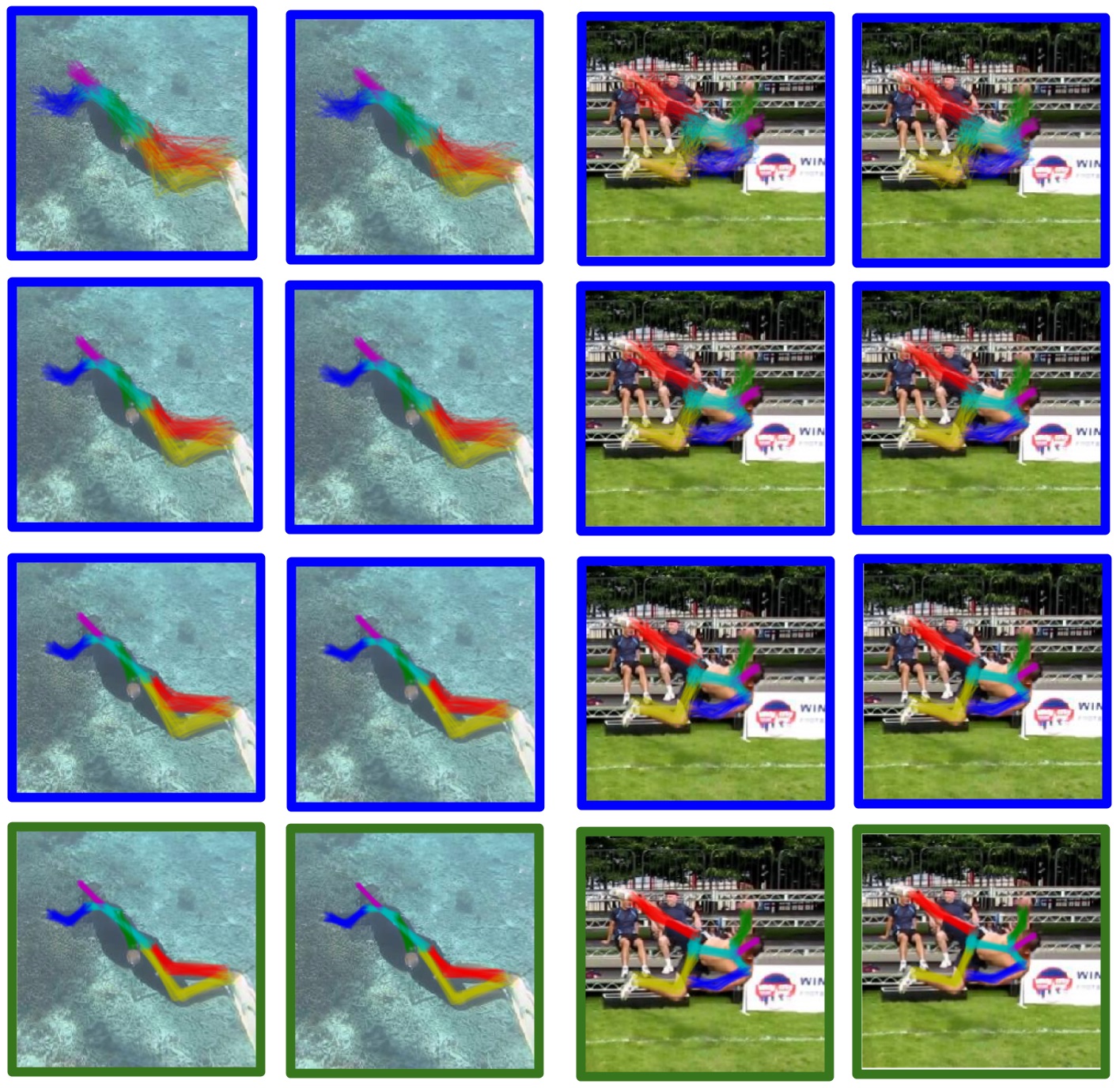}
	\end{center}
   	\caption{\em{Example of superimposed pose predictions by \textsc{Disco} Nets illustrating the uncertainty in the pose across training iterations for difficult examples. The blue box around the images represent a high diversity coefficient value, and the green box around them represents low diversity coefficient value. Columns $1$ and $3$ are outputs of the prediction network and columns $2$ and $4$ are outputs of conditional network. 	Row $1$ represents initial prediction of networks; rows 2 and 3 represents prediction of networks in second and fifth iteration respectively and last row represents prediction of networks when they have converged. The images in the first and second column show a rare action of person swimming underwater, and the third and fourth columns show a person in an unusual pose, where he is kicking the ball in air. Such rarity in pose leads to high uncertainty in both the networks initially. At convergence, both the networks provided accurate pose estimates for the difficult image by learning from the easier images. For each example, the first column shows estimated pose from prediction network and the second column shows estimated pose from conditional network. Best viewed in color.}}
    \label{fig:visualization-difficult}
\end{figure}

The final case, shown in figure \ref{fig:visualization-difficult} represents a difficult example, where the person is performing an unusual or rare action, like underwater swimming or a person kicking a ball in the air. The rarity of such poses in the supervised training set means that both prediction and conditional networks, $\Pr\nolimits_{\bf w}$ and $\Pr\nolimits_{\boldsymbol{\theta}}$, have high uncertainty in the predicted pose. However, over the iterations, by using the information gained from other simpler examples in the weakly supervised data set, the accuracy for such cases improves significantly. 

\clearpage

\section{Implementation Details}
\label{supsec:implementation_details}
In this section, we provide the details of our experimental setup. We construct $\Pr\nolimits_{\bf w}$ by taking a standard architecture for human pose estimation, namely, the stacked hourglass network \cite{newell2016stacked}. A noise filter of size $64 \times 64$ is added to the output of the penultimate hourglass module, which itself consists of $256$ $64 \times 64$ filters. The $257$ channels are convolved with a $1 \times 1$ filter to bring the number of channels back to $256$. This is followed by a final hourglass module as shown in figure~\ref{fig:disco-hg} (closely following stacking approach of Stacked Hourglass network \cite{newell2016stacked}. We note that all parameters remain differentiable and hence can be trained via backpropagation as discussed in Section~\ref{supsec:optimization} of the supplementary. 

The conditional network $\Pr\nolimits_{\boldsymbol{\theta}}$ is modeled exactly as the prediction network $\Pr\nolimits_{\bf w}$, except that there are ${\bf a}$ different output branches (consisting of $1$ hourglass module), one for each possible action class, stacked on top of penultimate hourglass module. Note that for each action class, we have a unique set of noise filters. During forward and backward propagation of the conditional network given an image from a particular action class, we mask the output from every other branch not corresponding to that particular action class. 

The non probabilistic pointwise network is  a {\sc Disco} Net that uses the architecture shown in figure~\ref{fig:disco-hg}, but discards the last two self-diversities terms in the learning objective (Equation~(\ref{eq:unbiasedObj})), and whose pointwise prediction is computed by principle of maximum expected utility ({\sc meu}) (Equation~(\ref{eq:h_predict})). We refer this pointwise network as {\sc PW} Net.

For the given data set ${\cal D}$, as given in section 4 of the paper, we train our three networks, {\sc FS}, $\text{\sc PW}_{\bf w}$ and $\Pr\nolimits_{\bf w}$ on the fully annotated training set. We note that after data augmentation, our training set (fully annotated data and the weakly annotated data) for each split, becomes $4\times$ larger, and for the {\sc FS} network, we additionally perform random crops such that the number of training samples for all three networks are the same. Networks $\text{\sc PW}_{\boldsymbol{\theta}}$ and $\Pr\nolimits_{\boldsymbol{\theta}}$ are first initialized by the weights of $\text{\sc PW}_{\bf w}$ and $\Pr\nolimits_{\bf w}$ respectively, then they are fine tuned using action specific samples from the fully annotated training set. For training, we used $\eta = 0.025$ and momentum $m = 0.9$. We cross validated weight decay regularization parameter $C$ in the range $[0.1, 0.01, 0.001, 0.0001]$ for our baseline networks {\sc FS} and {\sc PW} and found that values $0.001$ and $0.0001$ works best for  {\sc FS} and {\sc PW} respectively. We chose $C = 0.01$ for training our probabilistic networks. Moreover, for our probabilistic network, $\Pr\nolimits_{\bf w}$, we choose $K = 100$ samples. However, for a different task, it has been observed that results hold even for $K=2$ \cite{bouchacourt2016disco}.

While training the baseline non probabilistic point wise prediction network {\sc PW} using diverse data using self paced learning, we only backpropagate when the loss computed is within some threshold $t$. For such network, the loss would be high when predicted pose from $\text{\sc PW}_{\bf w}$ and $\text{\sc PW}_{\boldsymbol{\theta}}$ are very different from each other. Applying threshold on the loss for backpropagation ensures that these networks are only updated when both of them agree and therefore, they do not learn from erroneous or less confident predictions.

For our probabilistic network, $\Pr\nolimits_{\bf w}$, we do not require such threshold as the diversity coefficient term in our objective function ensures that our network learns only from confident predictions and not from samples when the network has low confidence. In other words, our method has fewer parameters than the baseline. 

We train all of these networks for 100 epochs and monitor the training and validation accuracies for each epoch. We employ an early stopping strategy based on validation accuracy to avoid over-fitting the data set. We save the network parameters corresponding to the best validation accuracy and report our result on the held out test set.


\section{Results}
In this section, we provide additional results of training the three network ({\sc FS}, {\sc PW} and $\Pr\nolimits_{\bf w}$) described in section~\ref{sec:experiments}.

\subsection{Results on MPII data set}

\begin{figure}[h]
	\begin{center}
    	\includegraphics[width=\linewidth]{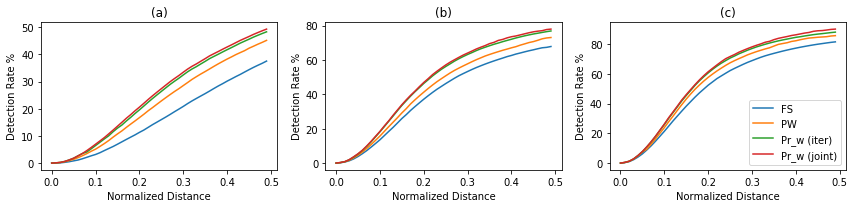}
    \end{center}
    \vspace{-5mm}
    \caption{\emph{Total PcKh comparison on MPII when trained on (a) $25-75$ split, (b) $50-50$ split; and (c) $75-25$ split.}}
	\label{fig:results}
\end{figure}

The detailed PcKh graphs on MPII data set by training an 8-stack hourglass network on various setting described in the paper are presented in figure~\ref{fig:results}.

In the figure, we can see that we consistently outperform the baseline {\sc FS} and {\sc PW} networks across all normalized distances. The networks trained on diverse data set (the {\sc PW} and the $\Pr\nolimits_{\bf w}$ network) performs significantly better on lower normalized scores than the {\sc FS} net which does not utilize the action annotations when there are only a few strong pose annotations available. This shows the utility of using action annotations when pose annotations are missing. The importance of using the probabilistic framework can be seen for lower normalized distance for all three splits, where the $\Pr\nolimits_{\bf w}$ network effectively captures the uncertainty present in the data set. We observe that as the number of supervised samples in our diverse data set increase, the accuracy of all the networks improves for smaller normalized distance. The joint training of the $\Pr\nolimits_{\bf w}$ network also improves the results over the iterative optimization of $\Pr\nolimits_{\bf w}$ network.

\subsection{Results on JHMDB data set}
In this subsection, we provide additional results of training our various models based on 8-stack hourglass network~\cite{newell2016stacked} on the JHMDB data set~\cite{Jhuang:ICCV:2013} for $50-50$ split.  

The JHMDB data set, which consists of $33183$ frames from $21$ action class, have $13$ annotated joint locations. We split the frames from each action class into $\{70, 15, 15\}\%$ training, validation and test sets, which corresponds to $22883$ frames in the training set, and $4150$ frames in the validation and the test set. To create a diverse data set with $50-50$ split, we randomly drop pose annotations from $50\%$ from the frames of the training set, similar to those described in Section~\ref{sec:experiments}. 

The result for training the {\sc FS}, {\sc PW} and $\Pr\nolimits_{\bf w}$ networks for the $50-50$ split on the JHMDB data set are summarized in table~\ref{table:result_jhmdb}.

\begin{table}[h!]
\centering
\begin{tabular}{c|c|c|c|c}
\hline
Method & {\sc FS} & {\sc PW} & $\Pr\nolimits_{\bf w}$ (iter) & $\Pr\nolimits_{\bf w}$ (joint) \\
\hline
Total Accuracy & 80.01 & 85.77 & 89.90 & 91.25 \\
\hline
\end{tabular}
\vspace{2mm}
\caption{\em Results on JHMDB data set ({\em PCKh@0.5}), where {\sc FS} is trained using $50\%$ percentage of fully annotated data and {\sc PW} and $\Pr\nolimits_w$ are trained on $50-50$ split of fully annotated and weakly annotated training data. Here {\sc FS} and {\sc PW} are the fully supervised and the pointwise networks respectively, and $\Pr\nolimits_w$ (iterative) and $\Pr\nolimits_w$ (joint) is our proposed probabilistic network trained with block
coordinate optimization and joint optimization respectively.}
\label{table:result_jhmdb}
\end{table}

We observe that the accuracies of the three networks ({\sc FS}, {\sc PW} and $\Pr\nolimits_{\bf w}$) holds similar trends as we had seen for the MPII data set.
\renewcommand*{\arraystretch}{1.2}
\begin{table}[!b]
\centering
\begin{tabular}{|l|l|l|}
\hline
Method              & MPII & JHMDB \\ \hline
{\sc FS}             & 41.89 & 54.31 \\ 
{\sc PW}            & 54.37 & 66.19 \\ \hline
$\Pr\nolimits_{\bf w}$ (iterative) & 56.09 & 71.02 \\
$\Pr\nolimits_{\bf w}$ (joint)     & \textbf{57.28} & \textbf{72.61} \\ \hline
\end{tabular}
\vspace{2mm}
\caption{\em Results on MPII Human Pose data set and JHMDB data set ({\em PCKh@0.5}), where {\sc FS} is trained using $50\%$ percentage of fully annotated data and {\sc PW} and $\Pr\nolimits_w$ are trained on $50-50$ split of fully annotated and weakly annotated training data. Here {\sc FS} and {\sc PW} are the fully supervised and the pointwise networks respectively, and $\Pr\nolimits_w$ (iterative) and $\Pr\nolimits_w$ (joint) is our proposed probabilistic network trained with block
coordinate optimization and joint optimization respectively.}
\label{table:additional_result}
\end{table}

\section{Additional Results}
To prove the generality of our method, we provide additional results using a different architecture, as proposed by Belagiannis {\em et al.}~\cite{belagiannis2015robust}. The authors pose the problem of estimating human poses as regression and propose to minimize a novel Tukey's biweight function as loss function for their ConvNet. They empirically show that their method outperforms the simple $L2$ loss. The point-wise architecture, consisting of five convolutional layers and two fully connected layers is modified to a {\sc Disco} Net as shown in the figure~\ref{fig:vasilis} below. A $1024$ dimensional noise vector, sampled from a uniform distribution, is appended to the flattened CNN features, before applying fully connected layers.
\begin{figure}[h!]
	\begin{center}
		\includegraphics[scale=0.37]{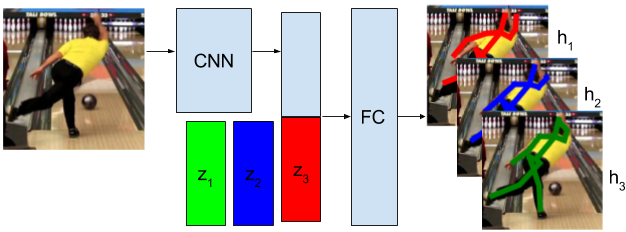}
	\end{center}
    \caption{Modified architecture, as proposed by Belagiannis {\em et al.}~\cite{belagiannis2015robust}. The figure shows the sampling process of {\sc Disco} Net. The block CNN consists of 5 convolution layers. The middle block is the flattened feature vector obtained after convolution. The block FC consists of two fully connected layers.} 
    \label{fig:vasilis}
\end{figure}

We evaluate the performance of the {\sc FS}, {\sc PW} and our proposed probabilistic network $\Pr\nolimits_{\bf w}$ on $50-50$ split of two data sets, namely (i) MPII Human Pose data set~\cite{andriluka14cvpr}, and (ii) JHMDB data set~\cite{Jhuang:ICCV:2013}. The various splits of MPII Human Pose are similar to the ones described in Section~\ref{sec:experiments}.  The MPII and the JHMDB data set is split exactly as it was done for the stacked hourglass network.
The results are summarized in Table~\ref{table:additional_result}.

We observe that the results shown in Table~\ref{table:additional_result} on both the data sets are consistent with our observations on the stacked hourglass network. Networks {\sc PW} and $\Pr\nolimits_{\bf w}$ trained on the diverse data, outperforms the {\sc FS} Net, which is trained only using the fully supervised annotations. This demonstrates the advantage of using diverse learning over a fully supervised method. Moreover, our proposed probabilistic net $\Pr\nolimits_{\bf w}$ outperforms the pointwise network {\sc PW}, this signifies the importance of modeling uncertainty over pose. We also note that performing joint optimization, after iterative optimization step, further increases our accuracy by $1.2\%$ on MPII Human Pose data set and by $1.4\%$ on JHMDB data set.

\end{document}